\documentclass{article}

\usepackage[preprint]{neurips_2023}
\usepackage[utf8]{inputenc} 
\usepackage[T1]{fontenc}    
\usepackage[dvipsnames]{xcolor}
\usepackage[colorlinks=true, urlcolor=Emerald ,linkcolor=Emerald,citecolor=Purple]{hyperref}
\usepackage{url}            
\usepackage{booktabs}       
\usepackage{amsfonts}       
\usepackage{nicefrac}       
\usepackage{svg}            
\usepackage{microtype}      
\usepackage{caption}
\usepackage{ragged2e}
\usepackage{multirow}
\usepackage{wrapfig}
\usepackage{xspace}
\usepackage{tcolorbox}
\usepackage{graphicx}
\usepackage{subcaption}
\usepackage{fancyhdr,lipsum}
\usepackage[export]{adjustbox}

\tcbuselibrary{most}

\title{TeenyTinyLlama: open-source \textit{tiny} language models trained in Brazilian Portuguese}

\author{                    
    Nicholas Kluge Corrêa$^{1,2}$
    \And
    Sophia Falk$^{1}$
    \And
    Shiza Fatimah$^{1}$
    \And
    Aniket Sen$^{1}$
    \And
    Nythamar de Oliveira$^{2}$
    \vspace{5mm}\\
    $^{1}$University of Bonn\\ 
    $^{2}$Pontifical Catholic University of Rio Grande do Sul\\
}

\begin{document}
\maketitle

\begin{figure}[h]
\centering
\includegraphics[width=0.45\linewidth]{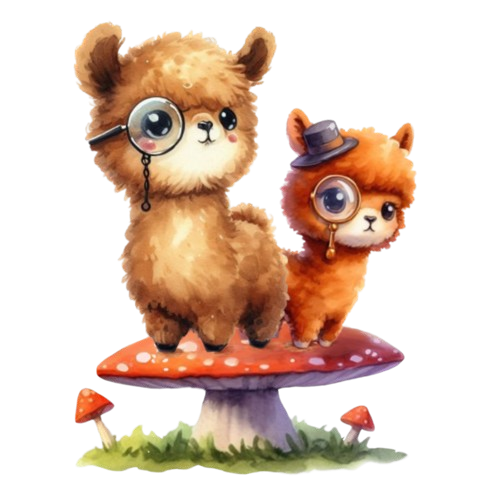}
\label{fig:TeenyTinyLlama-logo}
\end{figure}

\begin{abstract}

Large language models (LLMs) have significantly advanced natural language processing, but their progress has yet to be equal across languages. While most LLMs are trained in high-resource languages like English, multilingual models generally underperform monolingual ones. Additionally, aspects of their multilingual foundation sometimes restrict the byproducts they produce, like computational demands and licensing regimes. In this study, we document the development of open-foundation models tailored for use in low-resource settings, their limitations, and their benefits. This is the \textit{TeenyTinyLlama} pair: two compact models for Brazilian Portuguese text generation. We release them under the permissive Apache 2.0 license on \href{https://github.com/Nkluge-correa/TeenyTinyLlama}{GitHub} and \href{https://huggingface.co/collections/nicholasKluge/teenytinyllama-6582ea8129e72d1ea4d384f1}{Hugging Face} for community use and further development.

\end{abstract}

\section{Introduction}

Large language models have radically changed the field of natural language processing (NLP) with their exceptional ability to perform downstream tasks after being trained on vast amounts of data in a self-supervised learning regime. Under this paradigm, transformer-based models like BERT \citep{devlin2018bert}, RoBERTa \citep{liu2019roberta}, mT5 \citep{xue2020mt5}, and the whole family of GPT-style models \citep{radford2019language, black2022gpt, workshop2022bloom, biderman2023pythia, gunasekar2023textbooks, almazrouei2023falcon, touvron2023llama, luo2023yayi, jiang2024mixtral}, have become the foundation for many NLP applications and research areas. While part of the field is still pushing the search for new architectures, with innovations like the RWKV \citep{peng2023rwkv} and state space models \citep{gu2023mamba} promising new directions for research, the majority of LLM research is still focused on the scaling of model size, training data, and the general efficiency and capabilities of transformer-based LLMs \citep{vaswani2017attention}.

Despite the tremendous success of the field, progress has yet to be made equally regarding all languages. Hence, another trend in current NLP research involves the expansion of language domains with which such systems can interact. Current practices to tackle our linguistic multitude involve either training singular models in multiple languages \citep{conneau2019unsupervised, lin2021few, workshop2022bloom, shliazhko2022mgpt} or fine-tuning foundational models trained on multi-linguistic corpora to become monolingual or more proficient when working with low-resource languages \citep{eisenschlos2019multifit, pierre2020gpt2smallportuguese, alabi2022adapting, pires2023sabi, nguyen2023democratizing, lankford2023adaptmllm, zhao2024llama}.

However, most multilingual models available today still have a disproportional performance across languages due to the imbalance of training data, where usually high-resource languages, like English, represent the majority of such corpora, which creates user dissatisfaction with multilingual model's capabilities on non-English languages. Meanwhile, fine-tuned byproducts of multilingual models sometimes end up being restricted by the conditions imposed by the foundation used, like high computational costs for training and inference, which restrain adoption in low-resource settings, besides licensing regimes that prevent free use and open-source development. Factors like these highlight the necessity of building the foundations for monolingual LLMs for low-resource languages \citep{martin-etal-2020-camembert, souza2020bertimbau, scheible2020gottbert, antoun-etal-2021-aragpt2, nagoudi2021indt5, gutierrez2021maria, ko2023technical, rodrigues2023advancing}.

This study follows the trend of developing LLMs tailored for low-resource regimes \citep{gunasekar2023textbooks, zhang2024tinyllama, StableLM-2-1.6B}. In it, we sought to explore the challenges of developing LLMs in such settings. For this, we chose Brazilian Portuguese as our target language. To our knowledge, only a few LLMs for text generation were trained or fine-tuned to be proficient in Brazilian Portuguese and are available to the general public. Meanwhile, even fewer are available with permissive licenses and open-source code development. The models we developed, the \textit{TeenyTinyLlama} pair (TTL), were created to help democratize LLMs for low-resource languages and the open-source community in general, providing a simple and extensible implementation for LLM pre-training and fine-tuning at a small (< 2B parameters) scale.

\section{Related Works}

As already stated, multilingual models like BLOOM \citep{workshop2022bloom}, mGPT \citep{shliazhko2022mgpt}, and XGLM \citep{lin2021few} usually do not have satisfactory performance on low-resource languages, especially when compared to monolingual models. To overcome this, much of the community repurposes models trained on multilingual corpora to create mono-linguistic models for a target language via supervised fine-tuning (SFT), or, as some call it, extended pre-training \citep{pires2023sabi, larcher2023cabrita}.

Sometimes, this strategy even works on models not trained on multilingual datasets. For example, one of the first text-generation models for Brazilian Portuguese to appear to the general public was Pierre Guillou GPorTuguese-2 in 2020 \citep{pierre2020gpt2smallportuguese}, which coincided with the release of one of the first LLM natively trained in Brazilian Portuguese (BERTimbau) \citep{souza2020bertimbau}. GPorTuguese-2 is a byproduct of fine-tuning GPT-2 small \citep{radford2019language} on the Portuguese portion of the Wikipedia dataset \citep{wikidump} while also modifying the structure of the original model, giving it a new byte-pair encoding (BPE) tokenizer and repurposing the joint embeddings from the original model. While capable of generating fluent text in Brazilian Portuguese, GPorTuguese-2 fails to achieve a perplexity score on par with GPT-2 small, as documented by OpenAI, which is not surprising, given the limited hours of training ($\approx$ 30 hours of training) it received. Regardless, this pioneering work is available to all users under a permissive MIT License (as all GPT-2 models).

However, since the proposition of the GPT-2 architecture, many advances in transformer design have been made, and currently, more modern models are the default choice for engineers \citep{shoeybi2019megatron, black2022gpt, jiang2023mistral}. Perhaps one of the most significant contributions to the open source community in 2023 was the release of the Llama 2 architecture \citep{touvron2023llama}. Being the successor of Llama \citep{touvron2023llama1}, it brings many improvements that make training and inference of transformer-based language models more efficient, like the use of grouped query attention \citep{ainslie2023gqa}, better sub-layer normalization techniques \citep{zhang2019root}, changing ReLU activation's by SwiGLU \citep{shazeer2020glu}, and the use of rotary positional embeddings instead of positional ones \citep{su2021roformer}, besides being trained on a massive pre-training corpus, even beyond what is estimated to be optimal by the scaling Chinchilla laws \citep{hoffmann2022training}.

Currently, there is an entire ecosystem of Llama-based models being released on a Cambrian explosion rate \citep{openlm2023openllama, luo2023yayi, bi2024deepseek, zhang2024tinyllama, nous2024hermes, roziere2023code}, with other open-architectures like Mistral \citep{jiang2023mistral} also being heavily used by the open source community. Yet, it is from the LLama models that most of the other Brazilian Portuguese models come from. Three byproducts of fine-tuning (or extending the training of) Llama 1/2 with Brazilian Portuguese corpora are Bode \citep{bode2024}, Sabiá \citep{pires2023sabi}, and Canarim \citep{maicon_domingues_2023}.

Bode is a low-rank adaptation (LoRA) of Llama 2 fine-tuned with a translated version of the Alpaca dataset \citep{taori2023alpaca}, which contains 52K instruction-following demonstrations generated by text-davinci-003 \citep{ouyang2022training}. Bode is offered as LoRA adapters for Llama 7B and 13B while having a fine-tuned version of Llama 7B without using LoRA or other parameter-efficient fine-tuning techniques. In short, Bode only went through a fine-tuning phase, and the results of the model capacity are documented in their paper \citep{bode2024}. According to these, Bode stays within the performance of models that underwent a similar training regime, like Cabrita \citep{larcher2023cabrita}. It is also on par with its base model and even surpasses it on specific evaluations performed by the authors. However, like all derivatives of Llama models, Bode is licensed under the Llama 2 Community License Agreement, which is not as permissive as Apache 2.0, MIT, or commercial versions of the Creative Commons licenses.

Meanwhile, Sabiá models are fine-tuned versions of GPT-J \citep{gpt-j}, and Llama trained on a filtered portion of the ClueWeb 2022 dataset \citep{overwijk2022clueweb22}, which equates to 7.3–7.8 billion tokens, according to GPT-J and Llama tokenizers, respectively. The outcomes of this extended training process are Sabiá-7B, 65B (both derivatives of Llama), and Sabiá-J (using GPT-J as a base). According to the authors, their evaluations show that Sabiá-65B outperforms Llama 2-65B and GPT-3.5-turbo on the Portuguese Evaluation Tasks (Poeta)
benchmark (a set of tests gathered by the authors of the Sabiá paper). However, Sabiá-65B and Sabiá-J are not available to the public,\footnote{Even though GPT-J is available under an Apache 2.0 License.} while Sabiá-7B, just like the Bode models, was released with a Llama 2 license. Also, none of these models' training, evaluation, or fine-tuning source codes have been released to the community.

Regarding Canarim, there is not much to be said except that it is also a Llama 2 licensed model that underwent an extended training process on 16 billion tokens from a Portuguese subset of Common Crawl (2023) and was further fine-tuned on two datasets, one for instruction-tuning \citep{domingues2023dataset} and the other for open-ended question answering with a focus on the ENEM exams. Again, the models are also under the licensing regime of the Llama model's community license. For the interested reader, there are many other examples of these types of models on repositories like Hugging Face \citep{souza2024samba, henrique2024caramelo, henrique2024harpia, camaret2024harpia}.\footnote{That is, LLMs repurposed for other languages via full or LoRA/PEFT fine-tuning not accompanied by a paper or report.}

Finally, there are the Cabrita models \citep{larcher2023cabrita}. Unlike the work done by \citet{bode2024}, \citet{pires2023sabi}, and \citet{maicon_domingues_2023}, Cabrita is a byproduct of fine-tuning OpenLLaMA 3B \citep{openlm2023openllama}, which comes with an Apache 2.0 license. Also, Cabrita models have a modified tokenizer, unlike Sabiá, Bode, and Canarim, which directly repurposed the original Llama tokenizers. Cabrita extended training was performed on a subset of the mC4 dataset \citep{2019t5}, using $\approx$ 7 billion tokens. Cabrita 3B, like its base model, is available under an Apache 2.0 license.

To our knowledge, open-source LLMs for text generation pre-trained solely in Brazilian Portuguese are inexistent, nor have the above projects open-sourced their training and evaluation methods for reproducibility and further community development. At the same time, all models cited, based on billion-sized transformers, require non-trivial computational resources to use, adapt, and reproduce in low-resource settings. Hence, this study aimed to produce a pair of compact LLMs in Brazilian Portuguese by pre-training them from scratch (the TTL pair), tailored for (and produced by) a low-resource environment. The rest of this work documents our development, experiments, and results.

\section{Pre-Training}

In this section, we will describe how we designed the training of our TTL models. This study was performed with a closed budget of 500 USD, forcing us to make many developmental decisions to reduce costs and optimize our pre-training runs. This low-resource setting also influenced the size of our models that, as already shown by \citet{pires2023sabi}, can range from 9,000 to 80,000 USD when training billion parameter-sized models on a dataset similar to what we have gathered. Meanwhile, training models with trillion tokens, even in small settings, is beyond what we could finance and what is possible to accumulate with available Brazilian Portuguese text datasets. However, scaling down allowed us to choose a range of sizes where, according to scaling laws \citep{kaplan2020scaling, hoffmann2022training}, our limited budget was enough to pay for the computing necessary to pre-train our models and evaluate them.

\subsection{Sizing Up Models and Datasets}

While empirical evidence seems to point to the fact that existing scaling laws \citep{hoffmann2022training} may not provide accurate predictions in situations where smaller models are trained for more extended periods \citep{touvron2023llama, zhang2024tinyllama}, in this study, we choose to use the \citet{hoffmann2022training} scaling laws, like done by \citet{dey2023cerebras}, to estimate the size of models. Even though extrapolating such boundaries might benefit smaller models, we did not have the budget (or tokens) to sustain longer runs.

According to \citet{hoffmann2022training}, we can model language modeling loss, $L$, as a function of model size $N$ (the number of parameters) and training dataset size $D$ (the number of tokens):

$$L(N,D)=\frac{A}{N^\alpha}+\frac{B}{D^\beta}+E.$$

Where $A$ = 406.4, $B$ = 410.7, $E$ = 1.69, $\alpha$ = 0.32, and $\beta$ = 0.28 are parameters estimated by the authors after fitting a regression model to a dataset of 400 language model training runs. In their paper, \citet{hoffmann2022training} present estimations for dataset size for many model sizes. With 70B parameters, Chinchilla requires 1.4T tokens according to these laws, which equates to roughly 20 tokens per parameter. Based on this average, we estimated an optimal dataset size for two models: 3.5 and 9.5 billion tokens for 160 and 460 million parameter models, respectively. We considered these to be fair sizes for this project, given that we would be able to train them without requiring much computing while, at the same time, the token count was still within a manageable range, i.e., something we could gather by relying on open-source datasets.\footnote{We also experimented with extended training and embedding transplant, like in other works \citep{bode2024, pires2023sabi, maicon_domingues_2023}. In one of our initial explorations, we recycled the GPorTuguese-2 tokenizer and embedding layer, using them as a replacement for the original tokenizer and embedding weights (which are of the same dimension, i.e., 768) used by Pyhtia-160m \citep{biderman2023pythia}. We then performed a test training run of 100,000 steps using the same hyperparameters and settings later described in our work. In our experiments, the loss curves exhibit significant variance, with sudden increases in loss, suggesting a lack of smooth convergence during the first quarter of training. This has led us further into favoring the idea of pre-training a model from scratch, which, in the end, allowed us to train our TTL pair smoothly from beginning to end.}

\subsection{Pre-Training Dataset}

We consider English to be a high-resource language because, with datasets like the Pile \citep{gao2020pile}, RedPajama \citep{together2023redpajama}, the ROOTS corpus \citep{laurenccon2022bigscience}, the Stack \citep{kocetkov2022stack}, UltraChat \citep{ding2023enhancing}, etc., one can easily have access to trillions of high-quality, domain-specific tokens. Currently, most available tokens in Brazilian Portuguese come from datasets like BrWaC \citep{wagner2018brwac}, ClueWeb22 \citep{overwijk2022clueweb22}, Wikipedia \citep{wikidump}, OSCAR \citep{abadji2022towards}, and other byproducts of massive web crawling that require considerable filtering and pre-processing \citep{OrtizSuarezSagotRomary2019, ortiz-suarez-etal-2020-monolingual, wenzek-etal-2020-ccnet, conneau-etal-2020-unsupervised}.

Studies like the ones performed by \citet{xue2023repeat} and \citet{muennighoff2023scaling} explore the challenges of training LLMs under token crises, i.e., settings where the amount of data available is constrained. Both studies mainly focus on the downsides of repeating data during training runs, given that training language models with fresh data seem to have beneficial outcomes \citep{lee2021deduplicating}. According to \citet{muennighoff2023scaling}, under token crisis scenarios, training with up to 4 epochs of repeated data yields minor changes to loss compared to unique data. After this mark, increased repetition yields less performance, eventually decaying to zero. With this in mind, we aimed to build a dataset allowing training runs to be extended up and pass the optimal range without reaching the 4 epoch mark.

Hence, the first portion of our dataset comprises a concatenation of open-source Brazilian Portuguese datasets. These include: Wikipedia \citep{wikidump}, CulturaX  \citep{nguyen2023culturax}, OSCAR  \citep{OrtizSuarezSagotRomary2019, ortiz-suarez-etal-2020-monolingual, abadji2022towards}, Common Crawl  \citep{ wenzek-etal-2020-ccnet, conneau-etal-2020-unsupervised}, and ROOTS \citep{laurenccon2022bigscience} datasets. As a filtering step, we also utilized some of the filters used in \citet{rae2021scaling}, besides using a fine-tuned BERTimbau \citep{souza2020bertimbau} to exclude samples classified above a pre-defined toxicity threshold.\footnote{This model is available on Hugging Face: \url{https://huggingface.co/nicholasKluge/ToxicityModelPT}.} This first portion equates to 4.1 billion tokens that occupy approximately 50 GB of memory. We call this first portion Pt-Corpus.

The second portion of our dataset was inspired by the many studies that show that models fine-tuned with demonstrations of instruction-following behavior perform better in many downstream tasks \citet{askell2021general, ouyang2022training, bai2022training, chung2022scaling, touvron2023llama, shen2023mixture}, leading us to experiment with the inclusion of such type of data as part of a pre-training corpus. For this, we utilized the following datasets: Instruct-PTBR \citep{moro2024dataset}, Gpt4all-J \citep{moreira2024dataset}, Bactrian-X \citep{li2023bactrianx}, Dolly 15K \citep{DatabricksBlog2023DollyV2}, and CosmosQA \citep{huang-etal-2019-cosmos}, many of which are translated versions of English native datasets. The concatenation of both portions is what we call Pt-Corpus-Instruct. 60\% of this corpus is plain Brazilian Portuguese text (e.g., books, articles, blogs, etc.), while 40\% demonstrate instruction-following behavior. Pt-Corpus-Instruct equates to approximately 6.2 billion tokens that generate a memory footprint of 80 GB. Using \citet{muennighoff2023scaling} work, we estimated that this amount would be sufficient to train our models to pass the optimal point and up to 1 billion parameters if desired.

\subsection{Tokenization}

As pointed out by \citet{cui2023efficient} and \citet{larcher2023cabrita}, one of the obstacles related to adapting LLMs to new low-resource languages is the recycling of the tokenizer. For example, since Llama 2 was trained on a primarily English corpus, its tokenizer requires more tokens to encode non-English languages, shattering the information that could be already encoded into words or sub-units of words, as it does with much of the English vocabulary. However, one cannot merely exchange the tokenizer of a language model without some surgical adaptation of the embedding layer, as done by \citet{pierre2020gpt2smallportuguese}, given that this exchange would break the learned mapping between tokens and embeddings.

Since the model architecture we adopt in this study is Llama 2, we trained a Sentencepience tokenizer \citep{kudo2018sentencepiece} to make our model compatible with the ever-growing Llama ecosystem. We trained our tokenizer on 2 million text samples from our dataset, with a vocabulary size of 32K tokens. To test its efficiency, we performed the same test used by \citet{larcher2023cabrita} to access their tokenizer, where we counted the number of tokens required to encode 7400 words. According to the results in Table \ref{tab:tokenizer}, our tokenizer shows a 66\% improvement in efficiency compared to the original Llama 2 tokenizer, allowing for a more efficient way to encode Brazilian Portuguese text.

\begin{table}[htbp]
  \centering
  \normalsize
  \caption{ 
    Tokenizer Efficiency
  }
  \scalebox{1.0}{
  \begin{tabular}{lccc}
    \toprule
    \textbf{Model Tokenizer} & \textbf{nº of tokens} & \textbf{Vocabulary Size}\\
    \midrule    
    TTL & \textbf{9,937} & 32,000 \\
    GPorTuguese-2 & 9,959 & 50,257 \\
    BERTimbau & 11,006 & 29,794 \\
    Cabrita-3B & 11,488 & 52,000 \\
    Sabiá-7B & 14,813 & 32,000 \\
    \bottomrule
  \end{tabular}
  }
  \label{tab:tokenizer}
\end{table}

We utilized our tokenizer to encode our dataset into sequences of 2048 tokens. The raw text datasets and their tokenized versions are available for download on \href{https://huggingface.co/collections/nicholasKluge/teenytinyllama-6582ea8129e72d1ea4d384f1}{Hugging Face}.

\subsection{Architecture}

As done by \citet{zhang2024tinyllama}, we used a decoder-only Transformer model \citep{vaswani2017attention} based on Llama 2 \citep{touvron2023llama} as the basis for our models. The dimensions of our models are documented in Table \ref{tab:model}.

\begin{table}[htbp]
  \centering
  \normalsize
  \caption{ 
    TTL model's architecture
  }
  \scalebox{1.0}{
  \begin{tabular}{lcccccccc}
    \toprule
    \textbf{Size} & \textbf{Hidden size} & \textbf{Intermediate size} & \textbf{Context length}& \textbf{Heads} & \textbf{Layers} & \textbf{Vocab size}\\
    \midrule    
    160M & 768 & 3,072  & 2,048 & 12 & 12  & 32,000  \\
    460M & 1,024 & 4,096 & 2,048 & 16 & 24  & 32,000  \\
    \bottomrule
  \end{tabular}
  }
  \label{tab:model}
\end{table}

Our models have all the implementations that the Llama 2 architecture benefits from, i.e., grouped query attention \citep{ainslie2023gqa}, root mean square layer normalization \citep{zhang2019root}, SwiGLU activation's \citep{shazeer2020glu}, and RoPE embeddings \citep{su2021roformer}. While our 160 million parameter model uses 12 attention heads paired with 12 key-value heads, the 460 million parameter version uses 16 attention and key-value heads.

\subsection{Training}

We created all of our code implementations using the libraries tied to the Hugging Face ecosystem, i.e., Transformers \citep{wolf-etal-2020-transformers}, Datasets \citep{lhoest-etal-2021-datasets}, Tokenizers \citep{tokenizers}, and Accelerate \citep{accelerate}, which allow for easy reproducibility, adaptation, and further scaling. Our training and evaluation scripts follow a standard PyTorch structure \citep{paszke2019pytorch}, while we utilized CodeCarbon \citep{codecarbon} and Weights \& Biases \citep{wandb} for tracking our experiments.

Regarding hardware, we were limited to using a single NVIDIA A100-SXM4-40GB. To optimize its use, we performed several experiments to find the least costly training configuration regarding computing time and resource consumption, mainly regarding GPU memory utilization and tokens per second throughput. In these experiments, we explored the use of different mixed precision strategies and math modes (fp32, fp16, bf16, tf32), gradient accumulation steps, gradient checkpoints \citep{chen2016training}, the use of FlashAttention \citep{dao2022flashattention, dao2023flashattention}, different types of optimizers \citep{kingma2014adam, loshchilov2017decoupled, shazeer2018adafactor, dettmers2022optimizers}, and the use of data preloading versus data streaming. Ultimately, we arrived at the following training configurations, showcased in Table \ref{tab:model-config}, which produced a throughput of up to 29,491 tokens per second during training and 3$\times$ that during inference on an Ampere GPU. The hyperparameters related to the optimizer and learning rate scheduler were based on the documentation of other open-source LLMs of similar size \citep{zhang2022opt, workshop2022bloom, biderman2023pythia}.

\begin{table}[htbp]
    \centering
    \captionof{table}{TTL model's training configuration}
    \vspace{0.25cm}
        \label{tab:model-config}
    \begin{minipage}{0.5\linewidth}
        \centering
        TTL-160m \\
        \vspace{0.25cm}
        \begin{tabular}{lc}
            \hline
            Key & Value \\
            \hline
            tokens per batch & 8,192 \\
            total training steps & 458,000 \\
            gradient accumulation steps & 1 \\
            optimizer & AdamW \\
            learning rate & 6.0 $\times$ e$^{-4}$ \\
            adam epsilon & 1.0 $\times$ e$^{-8}$ \\
            adam beta 1 & 0.9 \\
            adam beta 2 & 0.999 \\
            weight decay & 0.01 \\
            scheduler type & cosine \\
            warmup steps & 5,000 \\
            gradient checkpointing & False \\
            mixed precision & bfloat16 \\
            tf32 & True \\
            flash attention 2 & True \\
            \hline
        \end{tabular}
    \end{minipage}%
    \begin{minipage}{0.5\linewidth}
        \centering
        TTL-460m \\
        \vspace{0.25cm}
        \begin{tabular}{lc}
            \hline
            Key & Value \\
            \hline
            tokens per batch & 8,192 \\
            total training steps & 1,200,000 \\
            gradient accumulation steps & 2 \\
            optimizer & AdamW \\
            learning rate & 3.0 $\times$ e$^{-4}$ \\
            adam epsilon & 1.0 $\times$ e$^{-8}$ \\
            adam beta 1 & 0.9 \\
            adam beta 2 & 0.999 \\
            weight decay & 0.01 \\
            scheduler type & cosine \\
            warmup steps & 10,000 \\
            gradient checkpointing & False \\
            mixed precision & bfloat16 \\
            tf32 & True \\
            flash attention 2 & True \\
            \hline
        \end{tabular}
    \end{minipage}
    
    \vspace{0.25cm}
    
The full details are available in our \href{https://github.com/Nkluge-correa/TeenyTinyLlama}{GitHub} repository.

\end{table}

This setting increased token throughput during training three times and in evaluation six times compared to using float32 precision, no tf32 mode, and a vanilla attention mechanism. The training of TTL-160m took approximately 36 hours (1.5 days), while the training of our 460 million parameter version took 280 hours (11.5 days).\footnote{Even though not used in this study, given the limitation of the number of GPUs we were able to use, our code implementation supports all parallel features from Accelerate, like distributed training on multiple GPUs, multiple nodes, and plugins to other distributed training libraries, like DeepSpeed \citep{rajbhandari2020zero}, FSDP \citep{ott2021fully}, and Megatron-LM \citep{shoeybi2019megatron}.}

During training, we saved several checkpoints for each model between an interval of 22,000 steps for TTL-160 and 25,000 for TTL-460m, resulting in 20 and 48 intermediate checkpoints, respectively. All checkpoints were saved along with the current state of their optimizer and scheduler, allowing our models to resume training at any checkpoint desired or for others to use these checkpoints as a starting point for further training or fine-tuning. At the same time, we measured, for each checkpoint, their estimated energy consumption and carbon emissions, which we used to compare with our model evaluation scores. We evaluated our models every 100,000 steps with a sample size corresponding to approximately 1\% of the training dataset. All models trained and checkpoints are available on \href{https://huggingface.co/collections/nicholasKluge/teenytinyllama-6582ea8129e72d1ea4d384f1}{Hugging Face}.

\section{Results}

\subsection{Evaluations}

\begin{figure}
    \centering
    \small
    \caption{Learning Curves for the TTL pair}
    \label{fig:training}
    \begin{subfigure}{0.42\textwidth}
        \centering
        \includegraphics[width = 200pt]{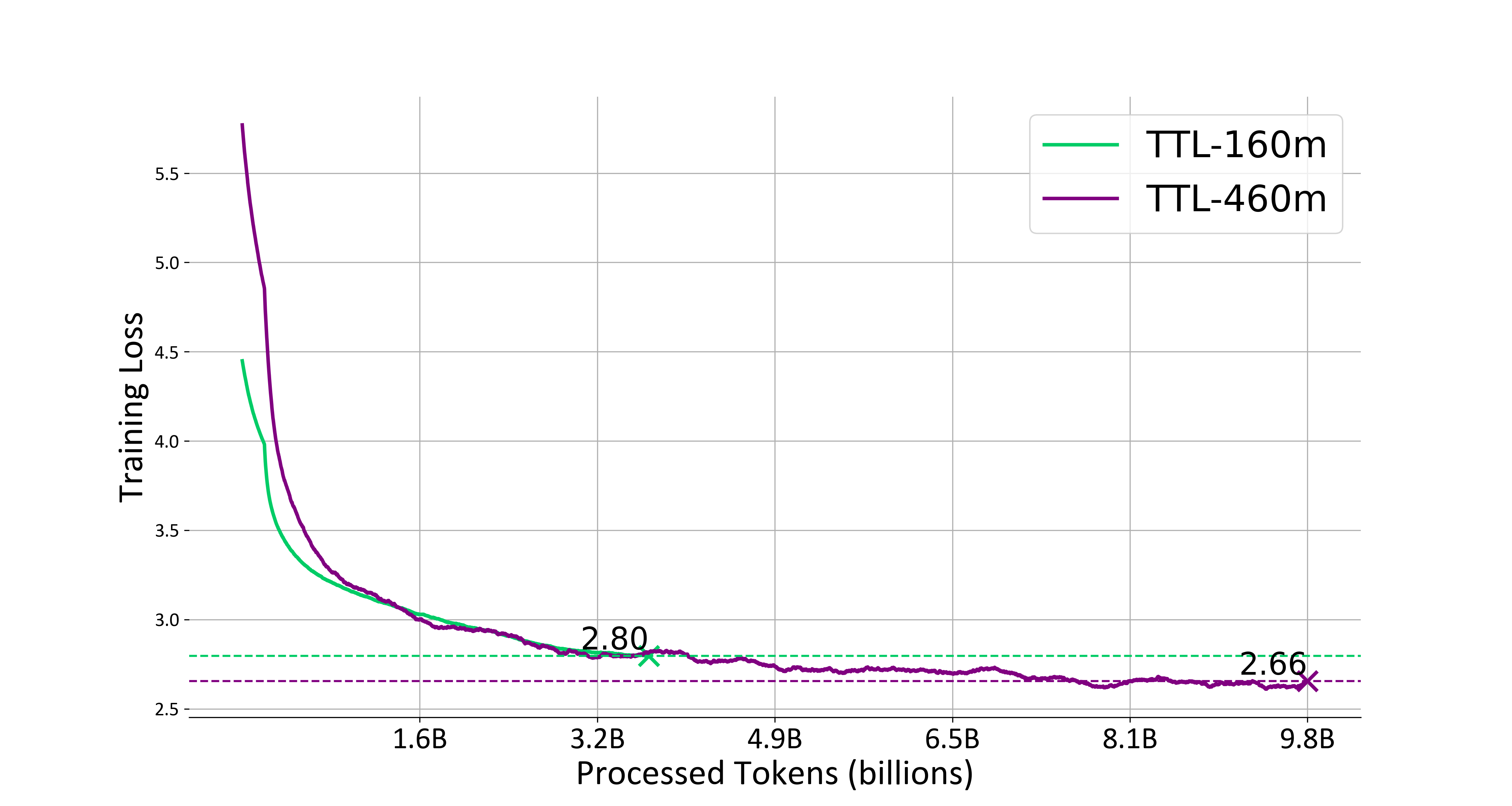}
        \caption{} 
        \label{fig:left}
    \end{subfigure}
    \begin{subfigure}{0.49\textwidth}
        \centering
        \includegraphics[width = 200pt]{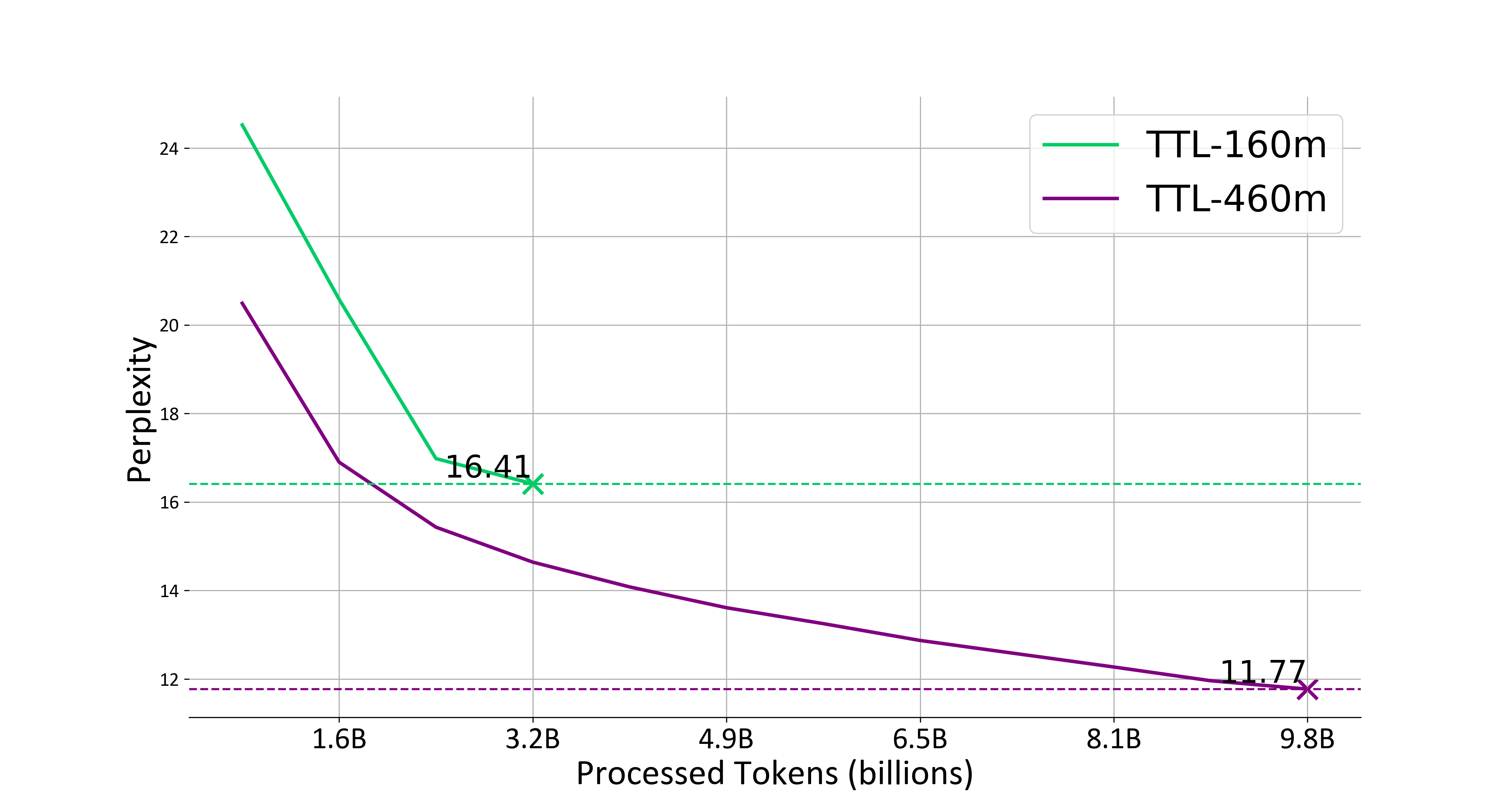}
        \caption{} 
        \label{fig:right}
    \end{subfigure}
    \vspace{0.25cm}

\justifying
\textbf{Figure 1:} Plot (a) shows the training loss of our TTL pair over their runs. TTL-460 was trained on 75,000 steps over our estimations (based on \citet{hoffmann2022training}. Both models show consistent convergence, and we speculate that they could be trained for longer with a significant increase in improvements, as demonstrated by \citet{biderman2023pythia}, on the training of even smaller models using over 300B tokens. Plot (b) shows the perplexity scores of our TTL pair measured at every 100,000 steps (8.1 million tokens). We were surprised that, even when trained on a comparatively smaller dataset, our models achieved a perplexity score similar to the results shown by \citet{radford2019language} for models of similar size. We attribute this to the superiority of the Llama 2 architecture compared to GPT-2, which, given the constant advances in the field, comes with several improvements that make training large neural networks more efficient.
\end{figure}

During our training runs, both models showed consistent convergence. At no point did our evaluation curves show signs of overfitting or saturation.\footnote{By saturation, we refer to the phenomenon where a model stops improving after a certain threshold of ingested tokens, probably due to the model size itself \citep{biderman2023pythia}.} In the case of our 460m parameter model, we intentionally trained past the optimal point by approximately 75,000 steps to assess if there were any signs of saturation, but our evaluations consistently gave better results. We hypothesize that our models are under-trained but can improve if further trained to pass the Chinchilla optimal range. As suggested by \citet{touvron2023llama} and  \citet{zhang2024tinyllama}, perhaps the scaling laws proposed \citet{hoffmann2022training} are indeed ill-suited to estimate the performance of small language models. In Figure \ref{fig:training}, we present the learning curves of our TTL pair.

Like many other low-resource languages, Brazilian Portuguese does not possess a standardized set of benchmarks that can be run to create model comparisons. Despite the introduction of the Poeta benchmark \citep{pires2023sabi}, as of the writing of this work, it still needs to come with a reproducible implementation of its use, which is a problem also reported by \citet{larcher2023cabrita}. On the other hand, benchmarks like the Language Model Evaluation Harness \citep{gao2021framework} permit a common way to test generative language models on few-shot evaluations. However, even though the evaluation harness possesses a plethora of benchmarks for testing, in its current implementation, these are not available for languages like Brazilian Portuguese.

To bypass these problems and perform an evaluation that can be reproduced and further expanded, we relied on the work of \citet{lai2023openllmbenchmark}, which translated four benchmarks from the original evaluation harness to 29 languages in a commendable effort. These are:

\begin{itemize}
    \item ARC-Challenge: a multiple-choice question-answering dataset containing questions from early grades science exams \citep{clark2018think}.
    \item HellaSwag: a multiple choice dataset for evaluating grounded commonsense inference \citep{zellers2019hellaswag}.
    \item MMLU: a benchmark that covers 57 subjects across STEM, humanities, social sciences, and more, measuring the performance of models on various natural language tasks \citep{hendrycks2020measuring}.
    \item TruthfulQA: a benchmark comprised of several questions, spanning 38 topics, that assess the model's tendency to replicate commonly believed falsehoods \citep{lin2021truthfulqa}.
\end{itemize}

\begin{figure}
    \centering
    \small
    \caption{TTL-460 accuracy on the ARC-Challenge during training}
    \label{fig:fine-tuning}
    \includegraphics[width = 400pt]{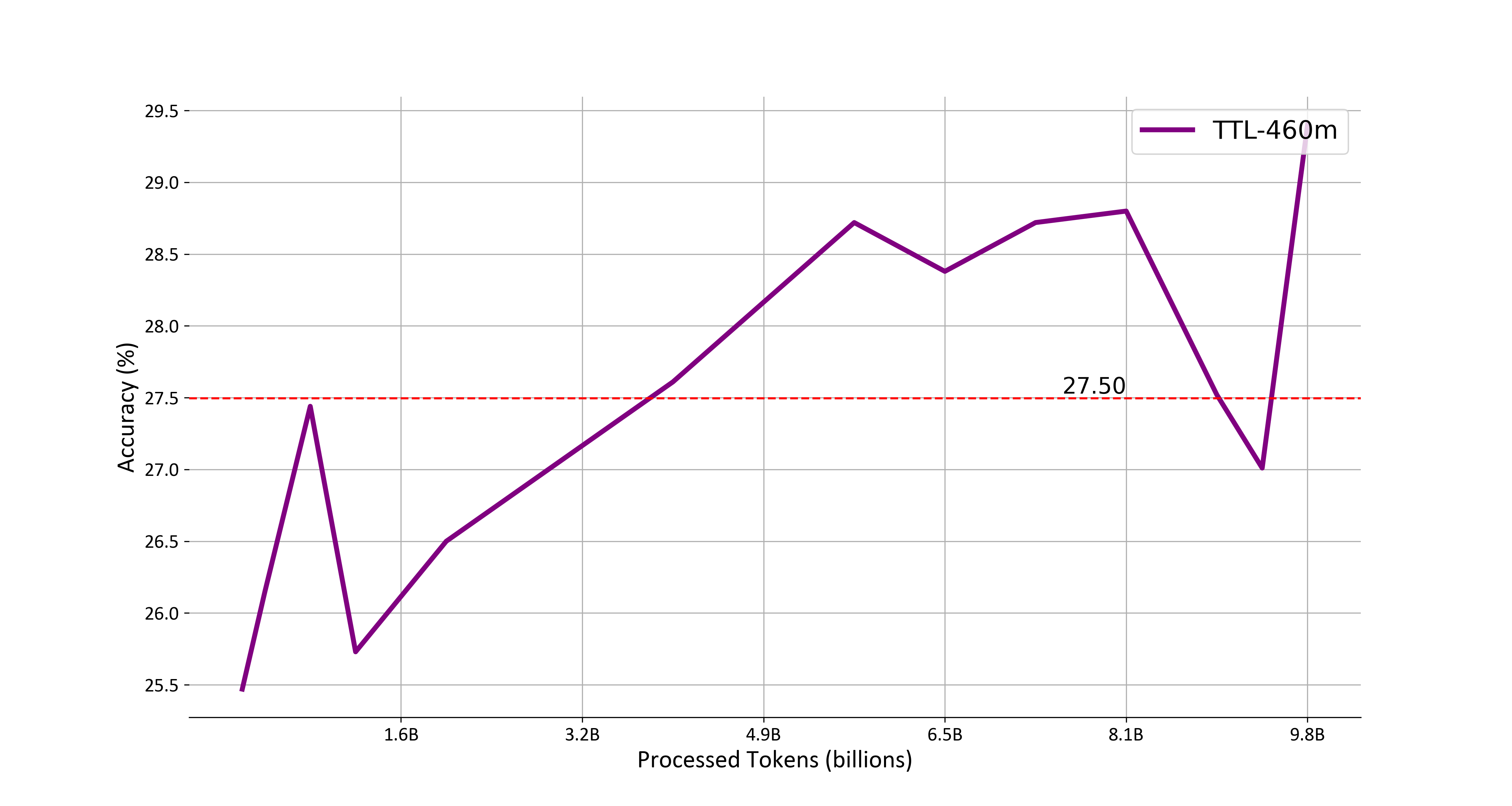}
    \vspace{0.25cm}
    
    \justifying
    \textbf{Figure 2:} According to the average accuracy score of TTL-460m on the ARC-Challenge, our model outperforms all other models listed in Table \ref{tab:benchmark-results}. Given our restricted budget, we could not evaluate all four benchmarks over our 68 checkpoints.
\end{figure}

For comparison purposes, we evaluated models categorized as within the same size range as the TTL pair on these benchmarks. Our results are in Table \ref{tab:benchmark-results}. Also, given that we saved several checkpoints of our models, this allows us to explore the evolution of LLMs capabilities as they are trained, as done by \citet{biderman2023pythia} and \citet{zhang2024tinyllama}. In Figure \ref{fig:fine-tuning}, we display the evolution of TTL-460m on the ARC-Challenge over 13 checkpoints.

\begin{table}[htbp]
  \centering
  \small
  \caption{Performance on the Language Model Evaluation Harness \citep{gao2021framework}.}
  \begin{tabular}{lccccccc}
    \toprule
    & \textbf{ARC} & \textbf{HellaSwag} & \textbf{MMLU} & \textbf{TruthfulQA} & \textbf{Average} \\
    \midrule 
    Pythia-410m & 24.83$^{*}$ & \textbf{41.29$^{*}$} & 25.99$^{*}$ & 40.95$^{*}$ & 33.26 \\
    \textbf{TTL-460m} & \textbf{29.40} & 33.00 & \textbf{28.55} & 41.10 & 33.01 \\
    Bloom-560m & 24.74$^{*}$ & 37.15$^{*}$ & 24.22$^{*}$ & 42.44$^{*}$ & 32.13 \\
    Xglm-564M & 25.56 & 34.64$^{*}$ & 25.18$^{*}$ & \textbf{42.53} & 31.97 \\
    OPT-350m & 23.55$^{*}$ & 36.73$^{*}$ & 26.02$^{*}$ & 40.83$^{*}$ & 31.78 \\
    \textbf{TTL-160m} & 26.15 & 29.29 & 28.11 & 41.12 & 31.16 \\
    Pythia-160m & 24.06$^{*}$ & 31.39$^{*}$ & 24.86$^{*}$ & 44.34$^{*}$ & 31.16 \\
    OPT-125m & 22.87$^{*}$ & 31.47$^{*}$ & 26.02$^{*}$ & 42.87$^{*}$ & 30.80 \\
    GPorTuguese-2 & 22.48 & 29.62 & 27.36 & 41.44 & 30.22 \\
    Gpt2-small & 21.48$^{*}$ & 31.60$^{*}$ & 25.79$^{*}$ & 40.65$^{*}$ & 29.97 \\
    Multilingual GPT & 23.81 & 26.37$^{*}$ & 25.17$^{*}$ & 39.62 & 28.73 \\
    \bottomrule
  \end{tabular}
  \label{tab:benchmark-results}
  
  \vspace{0.25cm}
  \justifying
  \textbf{Table 4:} All evaluations used the Language Model Evaluation Harness standard settings. Given our constrained budget, we could only evaluate some models in the Brazilian Portuguese version of the used benchmarks. Unfortunately, HellaSwag and the MMLU benchmarks require considerable time to run, even more so as the model size increases. Hence, results marked with an "*" were extracted from the Open LLM Leaderboard \citep{open-llm-leaderboard}, which uses the same evaluation method, using their English version. Thus, these results may vary if conducted in different languages, especially given that some models were pre-trained on mainly English text. Regardless, these results show that our models can perform as well and, in some instances (ARC and MMLU), surpass models trained on a much larger dataset and with many more resources. We speculate that this may result from our mixed dataset, which contains many demonstrations of instruction following and Q\&A.
\end{table}

To further evaluate the downstream capabilities of our models, we decided to employ a basic fine-tuning procedure of our TTL pair on a subset of tasks of the Poeta benchmark \citep{pires2023sabi}. Those are tasks involving toxicity detection \citep{vargas2022hatebr}, textual entailment \citep{chaves2023faquad, real2020assin}, sentiment analysis \citep{maas-EtAl:2011:ACL-HLT2011}, and a text classification \citep{Zhang2015CharacterlevelCN}. We apply the same procedure for comparison purposes on both BERTimbau models \citep{souza2020bertimbau}, given that they are also LLM trained from scratch in Brazilian Portuguese and have a similar size range to our models. Given their bidirectional nature, encoder-only transformers are usually superior in tasks like text classification, SQUAD-style Q\&A, and named entity recognition. However, we argue that we can still use these comparisons, especially if they are made in a standardized fashion, to assess if our pre-training runs produced LLM capable of producing good results ("good" here means "close to BERTimbau") when utilized for downstream applications. In this round of evaluations, we fine-tuned all models considered using the same setting and compared their final performance in Table \ref{tab:fine-tuning}.

\begin{table}[htbp]
  \centering
  \small
  \caption{Downstream performance on different tasks}
  \label{tab:fine-tuning}
  \begin{tabular}{lcccccc}
    \toprule
    \textbf{Models} & \textbf{IMDB} & \textbf{FaQuAD-NLI} & \textbf{HateBr} & \textbf{Assin2} & \textbf{AgNews} & \textbf{Average} \\

    \midrule
    BERTimbau-large & \textbf{93.58} & 92.26 & 91.57 & \textbf{88.97} & 94.11 & 92.10 \\
    BERTimbau-small & 92.22 & \textbf{93.07} & 91.28 & 87.45 & 94.19 & 91.64 \\
    \textbf{TTL-460m} & 91.64 & 91.18 & \textbf{92.28} & 86.43 & \textbf{94.42} & 91.19 \\
    \textbf{TTL-160m} & 91.14 & 90.00 & 90.71 & 85.78 & 94.05 & 90.34 \\

    \bottomrule
  \end{tabular}
  
  \vspace{0.25cm}
  \justifying
  \textbf{Table 5:} All the shown results are the higher accuracy scores achieved on the respective task test sets after fine-tuning the models on the training sets. All fine-tuning runs used the same hyperparameters: 3 epochs, batch size of 16, AdamW as the optimizer ($\alpha$ = 4e$^{-5}$, $\varepsilon$ = 1e$^{-8}$), and a weight decay rate of 0.01. Even though bidirectional encoder-only models usually perform better on the types of tasks under consideration, our larger model (TTL-410) can, in this simple fine-tuning setting, outperform BERTimbau-large on tasks involving toxicity detection and general text classification. Given that all results from our models present an over 90\% average accuracy score across tasks (even without pushing for optional hyperparameter settings), we argue that this shows the potential for our language models to be performative in many types of downstream tasks. All fine-tuned versions of TTL and their respective fine-tuning scripts are available on \href{https://huggingface.co/collections/nicholasKluge/teenytinyllama-6582ea8129e72d1ea4d384f1}{Hugging Face}.
\end{table}

We also measured the efficiency of our model in terms of its throughput capabilities. Given that generative language models in real-time applications are critically tied to their throughput and memory footprint, we estimated how many tokens our models can generate per second ($t/s$). According to our test, on average, TTL-460m can generate up to 12$_{t/s}$ on a Tesla V4 GPU. Applying a 4-bit activation-aware weight quantization \citep{lin2023awq} increases the throughput to 25$_{t/s}$, and reduces the model's memory footprint to 340 MB. On an A100, we increase throughput to 60$_{t/s}$, approximating 80 words generated per second. Further improvements can be achieved using inference frameworks that utilize more high-performance languages, like C or C++ (e.g., Llama.cpp).

Lastly, we licensed all models created by this project under an Apache 2.0 License.

\subsection{Energy Consumption and Carbon Emissions}

Given the consensus that tracking energy consumption, estimating carbon emissions, and reporting these results should be standard practice in the field of deep learning \citep{strubell2019energy, garcia2019estimation, lottick2019energy, lacoste2019quantifying, desislavov2021compute, luccioni2022estimating, falk2023challenging}, we logged our energy consumption during training and evaluation runs by using CodeCarbon \citep{codecarbon}. Besides achieving a count for the total energy consumption of our project, this allowed us to monitor performance and energy consumption increases coupled during a training run of an LLM. Table \ref{tab:energy-consumption} shows the logs for TTL-460m. According to it, performance improvements diminish midway through our training run while energy consumption and emissions rates remain constant. 

This observation underscores that nearly half of the energy consumed during our training runs corresponds to a marginal uptick in the model's performance. Meanwhile, this uptick becomes significantly tiny as the model approximates the optimal training point. Although it is unclear how long we can push training runs for smaller models, it is evident that the cost related to training large neural networks is directly proportional to model size, training time, and the hardware used.

\begin{table}[htbp]
  \centering
  \small
  \caption{Energy consumption during training (TTL-460m)}
  \begin{tabular}{cccccc}
    \toprule
    \textbf{Processed Tokens} & \textbf{Perplexity} & \textbf{Energy Consumption (kWh)} & \textbf{Emissions (KgCO2eq)} \\
    \midrule
    8.1M & 20.49 & 9.40 & 3.34 \\
    1.6B & 16.90 & 18.82 & 6.70 \\
    2.4B & 15.43 & 28.59 & 10.16 \\
    3.2B & 14.64 & 38.20 & 13.57 \\
    4.0B & 14.08 & 48.04 & 17.07 \\
    4.9B & 13.61 & 57.74 & 20.52 \\
    5.7B & 13.25 & 67.32 & 23.92 \\
    6.5B & 12.87 & 76.84 & 27.30 \\
    7.3B & 12.57 & 86.40 & 30.70 \\
    8.1B & 12.27 & 96.19 & 34.18 \\
    9.0B & 11.96 & 106.06 & 37.70 \\
    9.8B & 11.77 & 115.69 & 41.31 \\
    \bottomrule
  \end{tabular}
 
  \vspace{0.25cm}
  \justifying
  \textbf{Table 6:} Here, we display how the perplexity score of our model diminishes at every 100,000 steps (8.1 million tokens) and what the energy consumption (kWh) and estimated carbon emissions (CO2eq) are related to this process. As one can see, after half of our training run, our performance increase slows down as the rate of consumption and emissions keeps following a linear trend. This shows that, on our training runs, almost half of the energy we consumed was tied to a marginal increase in the model's performance ($\approx$ 1.84). By analyzing the training loss logs of other models \citep{black2022gpt, touvron2023llama, zhang2024tinyllama}, we argue that this is a common reality in the training of large-scale neural networks, i.e., \textit{convergence is slow and costly.}
  \label{tab:energy-consumption}
\end{table}

According to the estimations proposed by \citet{lottick2019energy}, implemented in CodeCarbon, in total, the 36 hours of compute time to train TTL-160m consumed 15.5 kWh ($\approx$ 5.7 KgCO2eq), while the 280 hours used to train TTL-460m consumed 113.0 kWh ($\approx$ 41.3 KgCO2eq). Summing all up, these emissions equate to a 185-kilometer car ride.\footnote{Calculations were made using the region of North Rhine-Westphalia (Germany) as the region of computing.} \footnote{GPU stats indicate that our training runs kept a steady allocation of GPU memory utilization (between 70\% - 85\% of its maximum capacity), power usage (83\%), and thermal output ($\approx$ 60º C).}

\subsection{Alignment}

With the release of ChatGPT in November 2022, there has been an increase in interest in models that went through an alignment process (e.g., instruction tuning, preference modeling, etc.), making them more attuned to follow the commands of people without the need for sophisticated prompting or further fine-tuning, becoming, in general, more helpful tools (a.k.a. assistants) to their users. Nowadays, there are many assistant models like ChatGPT \citep{nicholas22aira, taori2023alpaca, touvron2023llama, jiang2023mistral, koala_blogpost_2023, DatabricksBlog2023DollyV2, kopf2023openassistant}, which, besides being an object of interest to the general public, have become one of the most used laboratories for alignment research \citep{askell2021general}.

Our base models can follow instructions with minimal prompting, given that they were already exposed to millions of instructions during training. To further expand these capabilities, we fine-tuned the 460m parameter version of TTL on an instructional dataset to create a chat version of our larger base model, TTL-460m-Chat. Like in the Alpaca study \citep{taori2023alpaca}, we trained our chat model via SFT on a synthetically generated dataset. This dataset contains a collection of single-turn conversations between an assistant and a user, generated by prompting models that already went through an alignment process (ChatGPT, Vicuna, LLama 2, Open-Assistant, etc.). The dataset is available in Brazilian Portuguese and English and contains approximately 81K samples \citep{correa2023instructairadatasetv2}.\footnote{The Brazilian Portuguese version is a translated version of the English one, generated via the Google Translate API.}

\begin{figure}
    \centering
    \small
    \caption{Sample generated by TTL-460m-Chat}
    \label{fig:chat-sample}
    \includegraphics[width=400pt]{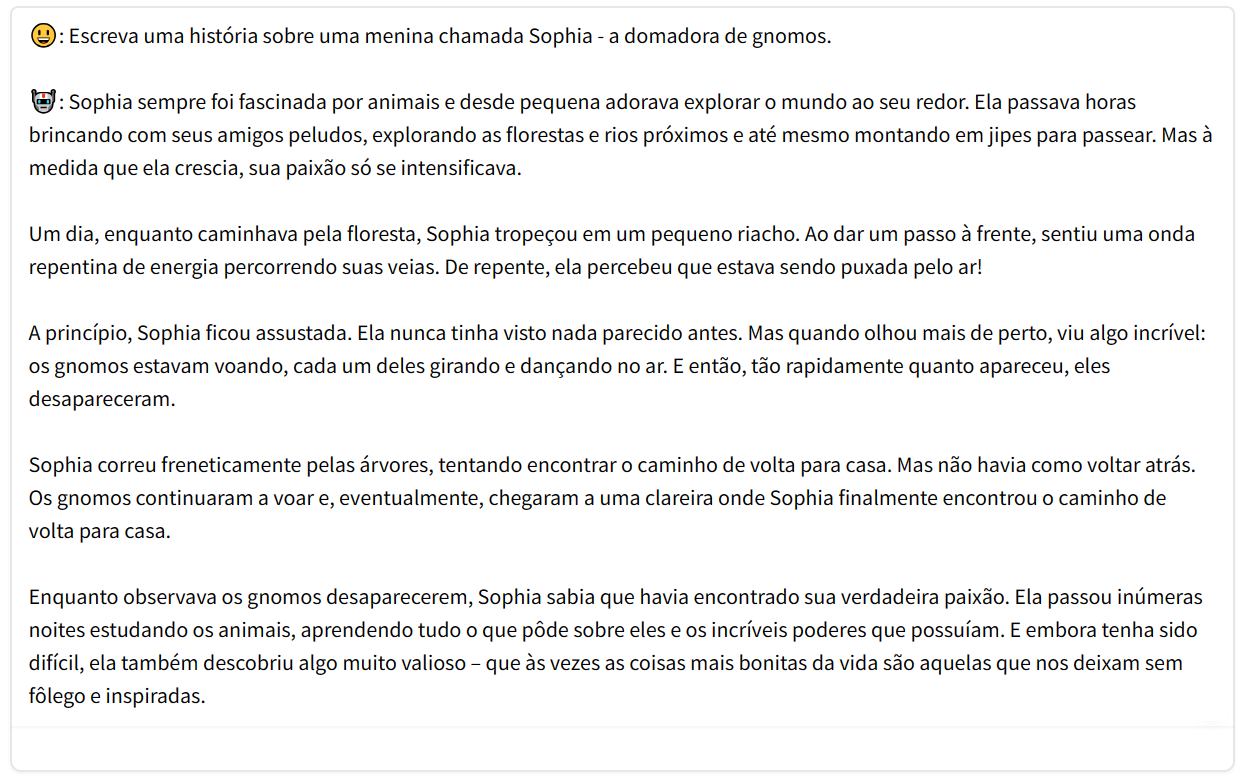}
    \vspace{0.25cm}
    
    \justifying
    \textbf{Figure 3:} Above is a user prompt requesting the model to generate a story; the English translation equates to: "\textit{Write a story about a girl called Sophia - the gnome tamer.}" Model Output: "\textit{Sophia has always been fascinated by animals, and from an early age, she loved exploring the world around her. She would spend hours playing with her furry friends, exploring the nearby forests and rivers, and even riding in jeeps for rides. But as she grew older, her passion only intensified. One day, while walking through the forest, Sophia stumbled across a small stream. As she stepped forward, she felt a sudden surge of energy coursing through her veins. Suddenly, she realized that she was being pulled through the air! At first, Sophia was frightened. She had never seen anything like it before. But when she looked closer, she saw something incredible: the gnomes were flying, each spinning and dancing in the air. And then, as quickly as they had appeared, they disappeared. Sophia ran frantically through the trees, trying to find her way back home. But there was no turning back. The gnomes continued to fly and eventually reached a clearing, where Sophia finally found her way home. As she watched the gnomes disappear, Sophia knew she had found her true passion. She spent countless nights studying the animals, learning everything about them and their incredible powers. And although it was difficult, she also discovered something precious - that sometimes the most beautiful things in life leave you breathless and inspired.}"
\end{figure}

For the SFT, we used the same software stack utilized to pre-train our models. TTL-460m-Chat was trained for three epochs using almost the same configurations documented in Table \ref{tab:model-config}. The only modifications are in the number of warm-up steps (1,000) and learning rate (1 $\times$ 10$^{-5}$). The full details are available in our \href{https://github.com/Nkluge-correa/TeenyTinyLlama}{GitHub} repository.\footnote{The additional training TTL-460m-Chat equated to an energy consumption of 5.6 kWh and an emission of 2.5 KgCO2eq. The procedure took approximately 13.5 hours.} We also created an open demo of our Chat model, which allows users to have conversations with TTL-460m-Chat.\footnote{Available in \url{https://huggingface.co/spaces/nicholasKluge/TeenyTinyLlama-Chat}.} In the demo, we also implemented a simple vector search engine that allows users to explore the fine-tuning dataset quickly, allowing them to evaluate the model's capabilities regarding "how much the model can go beyond its fine-tuning distribution." We also made available a 4-bit quantized version of this model for faster inference with almost no loss in performance. Figure \ref{fig:chat-sample} shows a sample of TTL-460m-Chat capabilities in story generation.

\section{Limitations}

Our work documents the process of developing text-generation language models for low-resource languages while being under a condition of low resources. With a budget of 500 USD, there is much that we could not perform, like multi-GPU distributed training with larger batch sizes and prolonged training runs. Even though our results indicate that we can achieve reasonable performance by limiting a training run to the stipulations made by specific scaling laws \citep{hoffmann2022training}, regardless, that evidence points out that our models are still under-trained. However, aware of the limitations of our work, we logged and documented every step in a reproducible manner, allowing others to push our work beyond what we currently can or begin new projects from our source. We still intend to run many tests and look forward to continuing to update this project.

Another aspect that limits this and every other work in language model development for low-resource languages is the need for more standard benchmarks to test the yields of our work. Machine learning is, to a great extent, a field moved by benchmarks. These tests allow us to define reasonable objectives and goals, work under a united evaluation framework, and compare our results in a standardized form. With this, progress becomes more accessible, and in this work, we tried to use evaluation methods that followed this motto. However, these are still limited, and there is a need to expand evaluation methodologies to encompass more languages. The field requires more work to translate or adapt current benchmarks to multiple languages. On the contrary, much of the research done in low-resource languages will continue to be (1) hard to reproduce and (2) difficult to compare.

Other limitations of this work are related to the TTL pair. Like almost all other language models trained on large text datasets scraped from the web, the TTL pair exhibited behavior that does not make them an out-of-the-box solution to many real-world applications, especially those requiring factual, reliable, nontoxic text generation. Our models are all subject to producing hallucinations, i.e., the generation of text that is incorrect, nonsensical, or not real, reproducing historical biases or generating toxic language, being overly verbose or repetitive, and generally producing content that should not be taken as given without human moderation.\footnote{A limitation of our models is how they generate code. Given that much of our training came from translated conversations on subjects like coding, our model tends to create code with programmatic commands that are natively written in English (\texttt{import torch}) on Brazilian Portuguese (\texttt{importar torch}), given the unstructured way in which people translated these samples.} Hence, even though our models are released with a permissive license, we urge users to perform their risk analysis on these models if intending to use them for real-world applications and also have humans moderating the outputs of these models in applications where they will interact with an audience, guaranteeing users are always aware they are interacting with a language model.

\section{Future Works}

The utility of language models, combined with the fact that their capabilities, for most languages, are still under-explored, presents many opportunities for researchers. In this work, we sought to document and expose the challenges of picking these low-hanging fruits while sharing our methods and tools with the community.

Regarding the TTL pair and LLM development on a smaller scale in general, we believe there is much room to be explored regarding their use and utilities, given that a model tailored for low-resource scenarios can be used for many applications, like edge computing, game development, real-time applications, and more. Besides, these sorts of models are necessary artifacts for research in general. Having models trained natively in the language of non-English speakers opens the doors to a whole population of non-English speakers who might be interested in research involving LLMs, be that on NLP, AI ethics, AI alignment, or other tangential fields.

With this in mind, here are some possible avenues for future projects seeking to expand our work:

\begin{enumerate}
    \item \textbf{Scaling to the 1B parameter range.} Accelerate permits a simple way to scale training on multiple GPUs, and at the 1B range, we are bound to encounter the emergence of improved capabilities \citep{gunasekar2023textbooks}. Given that our datasets, according to scaling laws for data-constrained models \citep{muennighoff2023scaling}, are sufficient to train models up to that range, future training runs on multiple GPUs might give us the first billion parameter text generation models trained on Brazilian Portuguese text.
    \item \textbf{Scaling dataset size to the 1T tokens mark.} To our knowledge, there are still no available trillion-sized datasets for Brazilian Portuguese. Collecting text datasets to create such a corpus would enable us to push the training of models to pass the Chinchilla scaling laws \citep{touvron2023llama1, touvron2023llama}, explore the saturation and training limits of smaller language models \citep{zhang2024tinyllama}.
    \item \textbf{Adding Brazilian Portuguese benchmarks to standard evaluation frameworks.} The more benchmarks we can bring to low-resource languages, the brighter future research for that language will be. Hence, adding tasks, new or translated versions, to evaluation frameworks like \citet{gao2021framework} language model evaluation harness is the type of grassroots work bound to benefit an entire research community. Also, testing all the Brazilian Portuguese language models mentioned in this work under a united framework would produce insightful results.
    \item \textbf{Expand the open-source development of language models for low-resource languages.} Our experiments could be reproduced for other languages if they can access a minimum amount of tokens. At the same time, open-source development should be encouraged \citep{kopf2023openassistant, openlm2023openllama, dey2023cerebras}, so in the future, the whole concept of "\textit{low-resource language}" will be a thing of the past.
\end{enumerate}

\section{Conclusion}

Large language models spearhead a paradigm shift in natural language processing, while widespread applications keep catalyzing these technologies to further push the field's boundaries. However, it is crucial to acknowledge that the advancements in LLMs have not been universally distributed across all languages and remain unevenly accessible, underscoring the ongoing challenge of achieving linguistic inclusivity openly and equitably.

In this study, our primary objective was to meticulously document the obstacles encountered and insights gained while training language models for low-resource languages, navigating the constraints imposed by a limited computational budget, a scarcity of available data, and a lack of standardized evaluation regimes.

Through this project, we successfully crafted a pair of language models, the \textit{TeenyTinyLlama} pair, trained in Brazilian Portuguese to the optimal range defined by the Chinchilla scaling laws. Remarkably, our findings indicate that these models exhibit comparable performance to other language models of similar size in various linguistic tasks while still demonstrating signs of possible undertraining.

All models, datasets, and source code developed in this study have been released under a permissive license, fostering open access and encouraging collaborative research within the academic community.

\section*{Acknowledgments}

This research was funded by RAIES (Rede de Inteligência Artificial Ética e Segura). RAIES is a project supported by FAPERGS (Fundação de Amparo à Pesquisa do Estado do Rio Grande do Sul) and CNPq (Conselho Nacional de Desenvolvimento Científico e Tecnológico).

\section*{Author's Information}

The corresponding author is \textbf{Nicholas Kluge Corrêa}. He is a postdoc researcher at the Center for Science and Thought at the University of Bonn (Bonn, NRW, Germany) and a fellow from the RAIES (Rede de Inteligência Artificial Ética e Segura) network (Porto Alegre, RS, Brazil). His contact email is \href{mailto:Nicholas.Correa@edu.pucrs.br}{Nicholas.Correa@edu.pucrs.br}.

\textbf{Sophia Falk} is a PhD researcher at the Bonn Sustainable AI Lab, Institute for Science and Ethics, University of Bonn. Her contact email is \href{mailto:falk@iwe.uni-bonn.de}{falk@iwe.uni-bonn.de}.

\textbf{Shiza Fatimah} is a master's student at the Institute for Computer Science, University of Bonn. Her contact email is \href{mailto:s39sfati@uni-bonn.de}{s39sfati@uni-bonn.de}.

\textbf{Aniket Sen} is a PhD researcher at the Helmholtz-Institut für Strahlen und Kernphysik, University of Bonn, and the Bethe Center for Theoretical Physics, University of Bonn. His contact email is \href{mailto:sen@hiskp.uni-bonn.de}{sen@hiskp.uni-bonn.de}.

Prof. Dr. \textbf{Nythamar de Oliveira} is a full professor at the Pontifical Catholic University of Rio Grande do Sul and the coordinator of the RAIES network. His contact email is \href{mailto:nythamar.oliveira@pucrs.br}{nythamar.oliveira@pucrs.br}.

\bibliographystyle{apalike}
\bibliography{references}

\end{document}